\documentclass[runningheads]{llncs}

\usepackage{eccv}

\usepackage{eccvabbrv}

\usepackage{graphicx}
\usepackage{booktabs}

\usepackage[accsupp]{axessibility}  %

\usepackage{hyperref}

\usepackage{orcidlink}

\usepackage{graphicx}
\usepackage{amsmath}
\usepackage{amssymb}
\usepackage{booktabs}
\usepackage{multirow}
\usepackage{rotating}
\usepackage[normalem]{ulem}
\useunder{\uline}{\ul}{}
\usepackage{epigraph} 
\usepackage{dsfont}
\usepackage{bm}
\usepackage{enumitem} %
\usepackage[accsupp]{axessibility}

\usepackage{pifont}%
\newcommand{\cmark}{\ding{51}}%
\newcommand{\xmark}{\ding{55}}%

\newcommand{\nerfmatch}{NeRFMatch\xspace}
\newcommand{\nerf}{NeRF\xspace}

\def\vec#1{\mathchoice%
	{\mbox{\boldmath $\displaystyle\bf#1$}}
	{\mbox{\boldmath $\textstyle\bf#1$}}
	{\mbox{\boldmath $\scriptstyle\bf#1$}}
	{\mbox{\boldmath $\scriptscriptstyle\bf#1$}}}

\definecolor{cvprblue}{rgb}{0.21,0.49,0.74}
\newcommand{\PAR}[1]{\noindent{\bf #1~}}

\begin{document}

\title{The NeRFect Match: Exploring NeRF Features for Visual Localization}

\titlerunning{NeRFMatch}

\author{Qunjie Zhou\inst{1}\textsuperscript{*}\orcidlink{0000-0002-2434-2393} \and
Maxim Maximov\inst{2}\textsuperscript{*}\orcidlink{0000-0002-2419-2815} \and
Or Litany\inst{1}\inst{3}\orcidlink{0000-0001-6700-7379} \and
Laura Leal-Taixé\inst{1}\orcidlink{0000-0001-8709-1133}}

\authorrunning{Q.~Zhou et al.}

\institute{
NVIDIA \and
Technical University of Munich, Germany \and
Technion, Israel
\\
\small{\texttt{
\href{https://nerfmatch.github.io}{https://nerfmatch.github.io}
}}
}

\maketitle

\def \thefootnote{*} 
\footnotetext{These authors contributed equally to this work}

\begin{abstract}
In this work, we propose the use of Neural Radiance Fields (\nerf) as a scene representation for visual localization. Recently, \nerf has been employed to enhance pose regression and scene coordinate regression models by augmenting the training database, providing auxiliary supervision through rendered images, or serving as an iterative refinement module. We extend its recognized advantages -- its ability to provide a compact scene representation with realistic appearances and accurate geometry -- by exploring the potential of \nerf's internal features in establishing precise 2D-3D matches for localization.
To this end, we conduct a comprehensive examination of \nerf's implicit knowledge, acquired through view synthesis, for matching under various conditions. This includes exploring different matching network architectures, extracting encoder features at multiple layers, and varying training configurations. Significantly, we introduce \nerfmatch, an advanced 2D-3D matching function that capitalizes on the internal knowledge of \nerf learned via view synthesis. 
Our evaluation of \nerfmatch on standard localization benchmarks, within a structure-based pipeline, achieves competitive results for localization performance on Cambridge Landmarks.
We will release all models and code.

\end{abstract}
  
\section{Introduction}
\label{sec:intro}

\begin{figure}[ht!]
\begin{center}
\includegraphics[width=0.85\linewidth, trim={0px 150px 60px 0px}, clip]{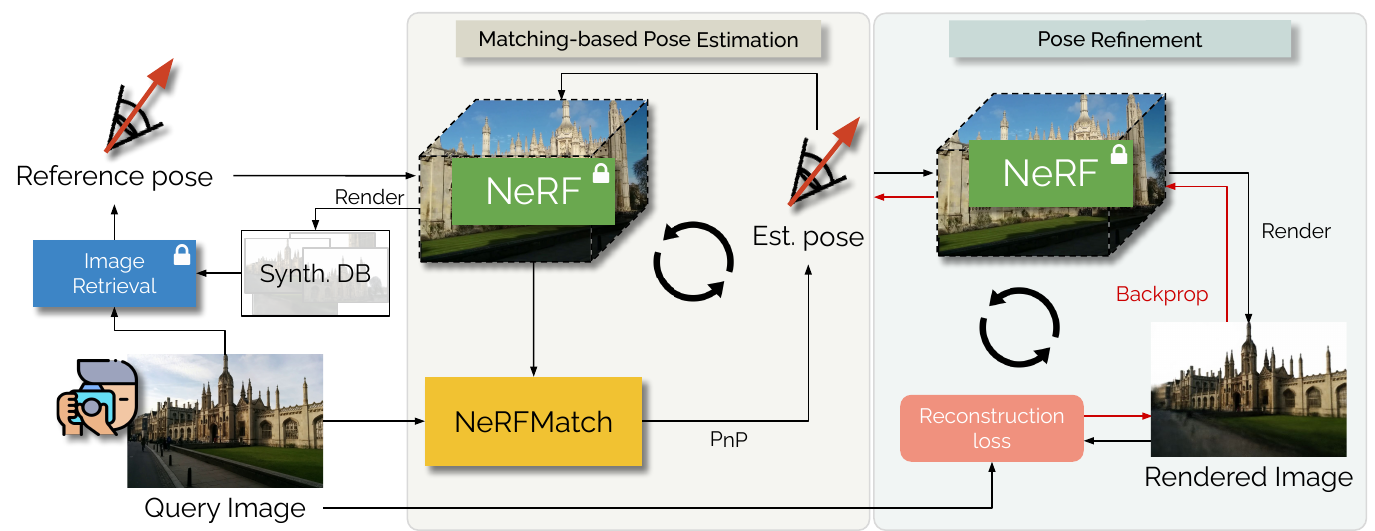}
\end{center}
   \caption{\textbf{\nerf-based localization overview.} In this work, we propose to use \nerf as our scene representation for visual localization. Given a query image, we first retrieve its nearest reference pose using image retrieval, then use \nerfmatch to establish 2D-3D correspondences between the query image and the \nerf scene points to compute an initial pose estimate and finally improve its accuracy via pose refinement. }
\label{fig:teaser}
\end{figure}

Visual localization is the task of determining the camera pose of a query image w.r.t a 3D environment. Such ability to localize an agent in 3D is fundamental to applications such as robot navigation~\cite{dong2009keyframe, wendel2011natural}, autonomous driving~\cite{heng2019autovision}, and augmented reality~\cite{arth2009wide, ventura2014global}. 
Different localization solutions can be categorized based on their underlying scene representation. Image retrieval~\cite{torii2015densevlad, hausler2021patchvlad, arandjelovic2016netvlad, berton2022cosplace} represents a scene as a database of reference images with known camera poses. 
A more compact representation utilizes a 3D point cloud, where each point is triangulated from its 2D projections in multiple views. Structure-based methods~\cite{taira2018inloc, li2010dubrovnik, sattler2016activesearch, irschara2009structure, sarlin2019hloc} rely on such a 3D model with associated keypoint descriptors to perform accurate localization. 
Recently, MeshLoc~\cite{panek2022meshloc} expanded structure-based localization by integrating dense 3D meshes, allowing to switch between different descriptors to establish the 2D-3D correspondences. 

Different from the aforementioned \textit{explicit} scene representations,  absolute pose regression (APR)~\cite{kendall2015posenet, walch2017poselstm, kendall2017geoposenet, brahmbhatt2018mapnet, blanton2020mspn, shavit2021mstpn} and scene coordinate regression (SCR) methods~\cite{shotton2013scene7, brachmann2017dsac, brachmann2018dsac++, brachmann2021dsac*, brachmann2023ace, li2020hierarchical, yang2019sanet} learn to encode scene information within network parameters, either directly in the localization network, or as a separate collection of latent codes as learned scene representation~\cite{tang2023neumap}. 
Despite being more compact than a sparse point cloud and a 3D mesh, learned \textit{implicit} scene representations are less interpretable and are limited to the task of visual localization. %

Recently, Neural Radiance Fields (NeRF)~\cite{mildenhall2020nerf} have emerged as a powerful representation of 3D scenes, encoded as continuous mapping of spatial coordinates and viewing angle to density and radiance. \nerf present several benefits: high interpretability, \ie, one can easily render scene appearance and depth at any given viewpoint, as well as being highly compact, \eg, a Mip-NeRF~\cite{barron2021mipnerf} model of $~5.28$ {MB} can represent a scene with a spatial extent ranging from  $1 m^2$~\cite{shotton2013scene7} to $~5 km^2$~\cite{kendall2015posenet} (\cref{sec:experiments}).
Owing to its attractive properties, \nerf is emerging as a \textit{prime} 3D scene representation~\cite{neuralfields}, alongside meshes, point clouds, and multi-view images, and has  been applied to various other computer vision tasks such as semantic segmentation~\cite{zhi2021semanticnerf, kundu2022panopticnerf, fu2022panoptic}, 3D object detection~\cite{xu2023nerfdet, hu2023nerfrpn}, Simultaneous Localization and Mapping (SLAM)~\cite{sucar2021imap, rosinol2022slamnerf, zhu2022niceslam, zhang2023go} and vision-based localization~\cite{yen2021inerf, chen2022dfnet, chen2023nerfscr, chen2023nefes, moreau2022lens, chen2021directpn, liu2023nerfloc, maggio2023locnerf_monte, moreau2023crossfire}.

\nerf has been leveraged for tackling visual localization in various ways. 
 iNeRF-style approaches~\cite{yen2021inerf, chen2023nefes} utilize a pre-trained \nerf as an inference-time pose refinement. Yet such methods commonly suffer from slow convergence and require pose initialization to be provided.
The end-to-end APR and SCR methods use a pre-trained \nerf only at train-time to augment training samples~\cite{moreau2022lens, chen2023nerfscr}, provide consistency supervision~\cite{chen2021directpn, chen2022dfnet}, and generate proxy depth ground-truth~\cite{chen2023nerfscr}. In this case, \nerf merely serves as an auxiliary representation.
Orthogonal to the above ways of leveraging \nerf, recent works~\cite{liu2023nerfloc, moreau2023crossfire} propose to use \nerf as a \textit{flexible} 3D model that can be enriched with volumetric descriptors to establish 2D-3D matches between an image and the scene. 
CrossFire~\cite{moreau2023crossfire} augments the base \nerf model with an additional 3D feature prediction branch. This feature is supervised to match 2D features extracted by an image backbone.
NeRF-Loc\cite{liu2023nerfloc} performs feature matching by utilizing a combination of features extracted from a generalizable \nerf and projected multi-view image features as 3D point features. 
However, both methods require training \nerf jointly with the matching task, prohibiting the usage of pre-built NeRF scenes.

In contrast, our work treats \nerf as the primary scene representation in visual localization without re-training and modifications.
Specifically, we focus on a crucial component -- \nerf features -- and 
(i) demonstrate their inherent capability in effectively supporting feature matching.
(ii) We introduce, \nerfmatch, a matching transformer that aligns 2D image features with 3D \nerf features and a minimal version of it to facilitate real-time applications. 
(iii) We present two options for performing pose refinement on top of \nerfmatch and conduct detailed analysis on their refinement effectiveness.
(iv) We use our \nerfmatch and pose refinement modules to perform hierarchical \nerf localization which achieves competitive localization performance on Cambridge Landmarks~\cite{kendall2015posenet}. Based on our experiments, we point out future work in needs to improve our indoor localization performance.  Our research paves the path towards localization  leveraging \nerf as the sole representation of the scene.

\section{Related work}
\label{sec:related_work}

\subsection{Visual Localization} 
\PAR{Structure-based localization.} Such methods~\cite{taira2018inloc, li2010dubrovnik, camposeco2019hsc, sattler2016activesearch, irschara2009structure, sarlin2019hloc} first estimate 2D-3D correspondences between a query image and the 3D points in the scene and then deploy a Perspective-n-Point (PnP) solver~\cite{ke2017p3p, gao2003p3p, kneip2011p3p} to compute the query camera pose. 
Image retrieval~\cite{torii2015densevlad, hausler2021patchvlad, arandjelovic2016netvlad, berton2022cosplace} is usually applied in advance to coarsely localize visible scene structure to a query image~\cite{sarlin2019hloc, taira2018inloc}.
Visual features~\cite{dusmanu2019d2net, detone2018superpoint, sarlin2020superglue, wang2020caps, zhou2021patch2pix, sun2021loftr, chen2022aspanformer} are often extracted from a database of scene images to represent 3D features and matched against the query image features extracted using the same algorithm to obtain 2D-3D matches. To optimize the inference runtime, 3D descriptors are cached with the scene  model at the cost of high storage demand and challenging map updates.
To relieve the burden coming from the need of storing visual descriptors, GoMatch~\cite{zhou2022gomatch} performs geometric feature matching, yet currently being less accuracy than visual feature matching approaches. 
Recently, to avoid storing massive amount of scene images as well as being flexible to switch between different descriptors for 2D-3D matching, MeshLoc~\cite{panek2022meshloc} propose to use 3D meshes as a dense 3D model. Compared to MeshLoc, we also pursuit a  dense scene representation via \nerf~\cite{mildenhall2020nerf}, which not only store scene images compactly but also provides free per-3D-point \nerf descriptors for direct 2D-to-3D matching.

\PAR{End-to-end Learned Localization.}
APR methods~\cite{kendall2015posenet, walch2017poselstm, kendall2017geoposenet, brahmbhatt2018mapnet, blanton2020mspn, shavit2021mstpn} directly regress a camera pose from a query image, while being lightweight by encoding scene information within a single model, they are currently less accurate than approaches based on 2D-3D matching~\cite{sattler2019apr}. 
In contrast, SCR methods~\cite{shotton2013scene7, brachmann2017dsac, brachmann2018dsac++, brachmann2021dsac*, brachmann2023ace, li2020hierarchical, yang2019sanet} perform implicit 2D-3D matching via directly regressing 3D scene coordinates from a query image. Similar to APR, they learn to encode the scene geometry within their own network parameters~\cite{brachmann2017dsac, brachmann2018dsac++, brachmann2021dsac*, brachmann2023ace, li2020hierarchical} but they are limited by the model capacity to memorize large-scale scenes. Recently, scene agnostic SCR methods~\cite{yang2019sanet, tang2023neumap} have been proposed to scale up to larger scenes by decoupling the scene representation from the learned matching function.

\subsection{\nerf in Localization}

iNeRF~\cite{yen2021inerf} directly inverts a \nerf model to refine a camera pose initialization by iteratively optimizing the photometric difference between the rendered image and the query image. However, it requires hundreds of iterations to  converge and thus is not directly applicable to real-world visual localization.
LENS~\cite{moreau2022lens} leverages \nerf as a novel view synthesizer (NVS) to augment the image database for pose regression training. Similar to LENS~\cite{moreau2022lens}, NeRF-SCR~\cite{chen2023nerfscr} augments SCR training samples using  RGB-D images rendered from a \nerf model based on an uncertainty-guided novel view selection. 
DirectPN~\cite{chen2021directpn} incorporates \nerf to provide photometric consistency supervision for pose regression where it minimizes the color difference between the query image and the image rendered at a predicted pose. DFNet~\cite{chen2022dfnet} extends this idea to measure the consistency in the feature space showing boosted localization performance. 
NeFes~\cite{chen2023nefes} follows iNeRF to use \nerf as an offline pose refinement module on top of DFNet. It distills pre-trained DFNet feature into a \nerf model and directly renders it for pose optimization.
Different from NeFes, we directly use 3D viewpoint-invariant feature learned during a standard \nerf training and validating its potential to deliver highly-accurate localization performance. 
In addition, our model is able to perform localization in a scene-agnostic (multi-scene) setting while DFNet and NeFes is scene-dependent.

The most relevant works to ours are NeRFLoc~\cite{liu2023nerfloc} and CrossFire~\cite{moreau2023crossfire} as they also establish explicit 2D-3D matches with features rendered from \nerf. 
NeRFLoc proposes a generalizable \nerf that is conditioned on a set of reference images and reference depths to output descriptors for 3D points by fusing multi-view image features, while CrossFire lifts an instant-NGP~\cite{muller2022instantnerf} model to directly outputs feature descriptors for 3D points.
In contrast to their methods that both require to train their customized scene model together with the matching model.
our \nerfmatch directly learn to align image feature with pre-trained \nerf features, which allows us to directly benefit from the on-going advancement in the typical \nerf research.

\section{\nerf-based Localization}
\label{sec:method}

In this work we explore the capability of  \nerf features, tasked with view synthesis, to offer precise 2D-3D correspondences for addressing visual localization. 
To this end, we introduce our \nerf-based localization pipeline, which adheres to the general steps of standard structure-based localization approaches ~\cite{sarlin2019hloc, taira2018inloc}. We provide an overview of the localization pipeline in \cref{sec:method_loc_pipeline}, followed by an explanation on how we utilize \nerf as a scene representation for localization in \cref{sec:method_nerf_scene}. 
Afterwards, we detail our iterative pose refinement component in \cref{sec:method_pose_refinement}.
Finally, we delve into the specific challenge of matching images to \nerf features using our newly proposed \nerfmatch model in  \cref{sec:method_nerfmatch}.

\subsection{Localization Pipeline}\label{sec:method_loc_pipeline}
Given a query image and a scene represented by a pre-trained \nerf model associated with a list of pre-cached reference poses, \eg, the poses used to train the \nerf, our localization pipeline involves three steps to localize the query as shown in \cref{fig:teaser}.
(i) We first apply image retrieval~\cite{arandjelovic2016netvlad, torii2015densevlad}, which extracts visual descriptors from the query image and the reference images (synthesized by \nerf at all reference poses, \cf \cref{exp:localization}) to find the nearest neighbour to the query based on descriptor distances. 
This step efficiently narrows down the pose search space to the vicinity of the reference pose.
(ii) We use the reference pose to render a set of 3D points with associated \nerf descriptors (\cf \cref{sec:method_nerf_scene}) and feed it together with the query image into
our \nerfmatch (\cf \cref{sec:method_nerfmatch}) to predict the 2D-3D matches between the query image and 3D points, from which we estimate the absolute camera pose of the query image via a PnP solver~\cite{ke2017p3p, gao2003p3p, kneip2011p3p}.
(iii) We further use our pose refinement module (\cf \cref{sec:method_pose_refinement}) to improve the pose estimation iteratively. Note that this step is optional considering the trade-off between accuracy and runtime efficiency.

\subsection{\nerf for Localization}\label{sec:method_nerf_scene} 

\begin{figure}[ht!]
\begin{center}
\includegraphics[width=0.8\linewidth, trim={0px 260px 0px 0px}, clip]{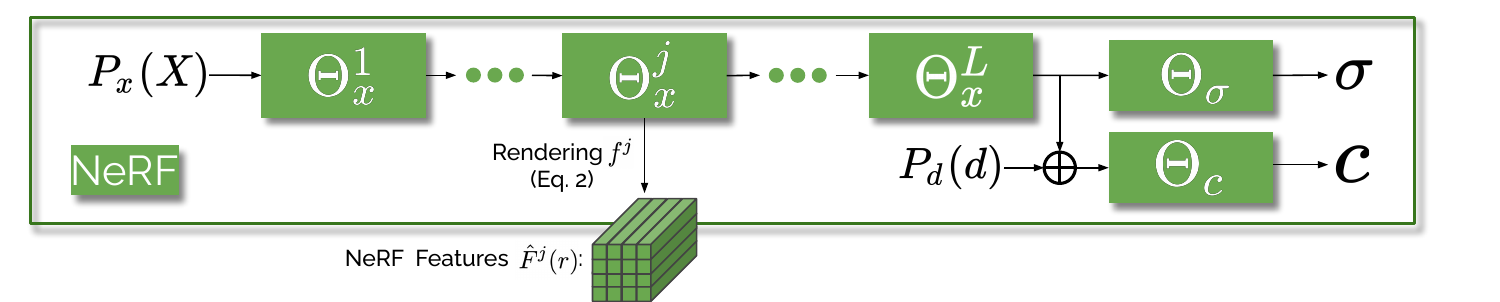}
\end{center}
   \caption{\textbf{An overview of the standard NeRF architecture.} The input consists of a scene coordinate $X$ and ray directions $d$. The outputs include color $c$, density $\sigma$. 
   We obtain intermediate features, denoted as $f^{j}$, using volumetric rendering. 
   }
\label{fig:nerf_arch}
\end{figure}

\PAR{\nerf architecture.}
\nerf~\cite{mildenhall2020nerf}  models a 3D scene with a coordinate network. Specifically, it maps a 3D point and a camera viewing direction to their corresponding volume density and RGB values. 
As depicted in \cref{fig:nerf_arch}, a standard \nerf model consists of two non-learnable positional encodings, $P_x$ for 3D coordinates and $P_d$ for viewing directions, and three trainable components: a 3D point encoder $\Theta_{x}$, a volume density decoder $\Theta_{\sigma}$, and a RGB decoder $\Theta_{c}$.
More concretely, the 3D encoder consists of $L$ layers, \ie, 
$\Theta_{x} = \Theta^{L}_{x} \circ \dots \circ \Theta^{1}_{x}$.
In this work, we are particularly interested in exploring the potential of 3D features extracted within the 3D encoder. Given a 3D point $X \in R^{3}$, we define its 3D feature extracted at $j$-th 3D encoder layer as $f^{j} = \Theta^{j}_{x} \circ \dots \circ \Theta^{1}_{x} (P_x(X))$, where $\Theta^{j}_{x}$ is the $j$-th layer in the 3D encoder.
The last layer 3D feature is next input into the density decoder to predict the volume density $\sigma = \Theta_{\sigma}(f^{L})$, while the color decoder takes a positional-embedded viewing direction $d \in R^2$ in addition to the 3D feature $f^{L}$ to compute the view-dependent color $c=\Theta_{c}(f^{L}, P_d(d))$. 

\PAR{Volumetric rendering of color.}
\nerf employs the continuous scene representation to render per-ray color using a discretized volumetric rendering procedure.
Given a ray $r(t) = o + td$  emitted from the camera center $o \in R^3$, along a viewing direction $d$ which intersects the image plane at pixel $x$, this ray is sampled at $N$ 3D points between a near and far planes. 
The RGB color $\hat{C}(r)$ at pixel $x$ is calculated as:  
\begin{equation}\label{eq:render_rgb}
\hat{C}(r) = \sum_{i=1}^{N} w_i c_i, \indent w_i = T_i (1-exp(-\delta_i \sigma_i)), \
\end{equation}
where $\delta=t_{i+1}-t_{i}$ is the sampling interval, $c_i$ and $\sigma_i$ are the predicted color and density of $i$-th sampled 3D point, and $T_i = exp(-\sum_{j=1}^{i-1} \sigma_j \delta_j )$~\cite{mildenhall2020nerf}.

\PAR{Volumetric rendering of 3D points and features.} 
We investigate the capability of \nerf features to act as descriptors for 3D surface points, facilitating 2D-3D correspondence with a query image. 
Following the volumetric rendering process defined in \cref{eq:render_rgb}, we define a rendered 3D (surface) point $\hat{X}(r)$ and  its associated \nerf descriptor $\hat{F}^j(r)$ along the ray $r$ cast through image pixel $x$ as the weighted sum of sampled 3D points and their respective features:
\begin{equation}\label{eq:render_3d}
\hat{X}(r) = \sum_{i=1}^{N} w_i X_i ,\indent \hat{F}^{j}(r) = \sum_{i=1}^{N} w_i f^j_i .
\end{equation}
Here, $X_i$ is $i$-th sampled 3D points, $f^j_i$ is its 3D feature extracted at $j$-th 3D encoder layer and 
$w_i$ is the weight computed from its density prediction~\cref{eq:render_rgb}.

\subsection{Pose Refinement}\label{sec:method_pose_refinement}
Following the matching-based pose estimation, we employ two approaches to refine the estimated camera pose. 
The first approach, iterative refinement, uses the estimated camera pose as a new reference for extracting \nerf features. This process involves repeating the matching procedure with the updated reference pose, incrementally enhancing the results due to the closer proximity of the reference pose, which makes the \nerf and image features more similar. While this process can be repeated multiple times, significant improvements are typically observed after the first refinement (\cf \cref{exp:pose_refinement}).
Inspired by iNeRF~\cite{yen2021inerf}, our second refinement option combines optimization and matching where we backpropagate through the frozen \nerf model to optimize an initial camera pose estimate by minimizing the photometric difference between the query image and the current \nerf rendered image. 
We then use the optimized camera pose to render again 3D points and features to perform the 2D-3D matching and compute the final refined camera pose.

\section{Image-to-\nerf Matching Network}\label{sec:method_nerfmatch}
\begin{figure}[t!]
\begin{center}
\includegraphics[width=.9\textwidth, trim={0px 133px 0px 0px}, clip]{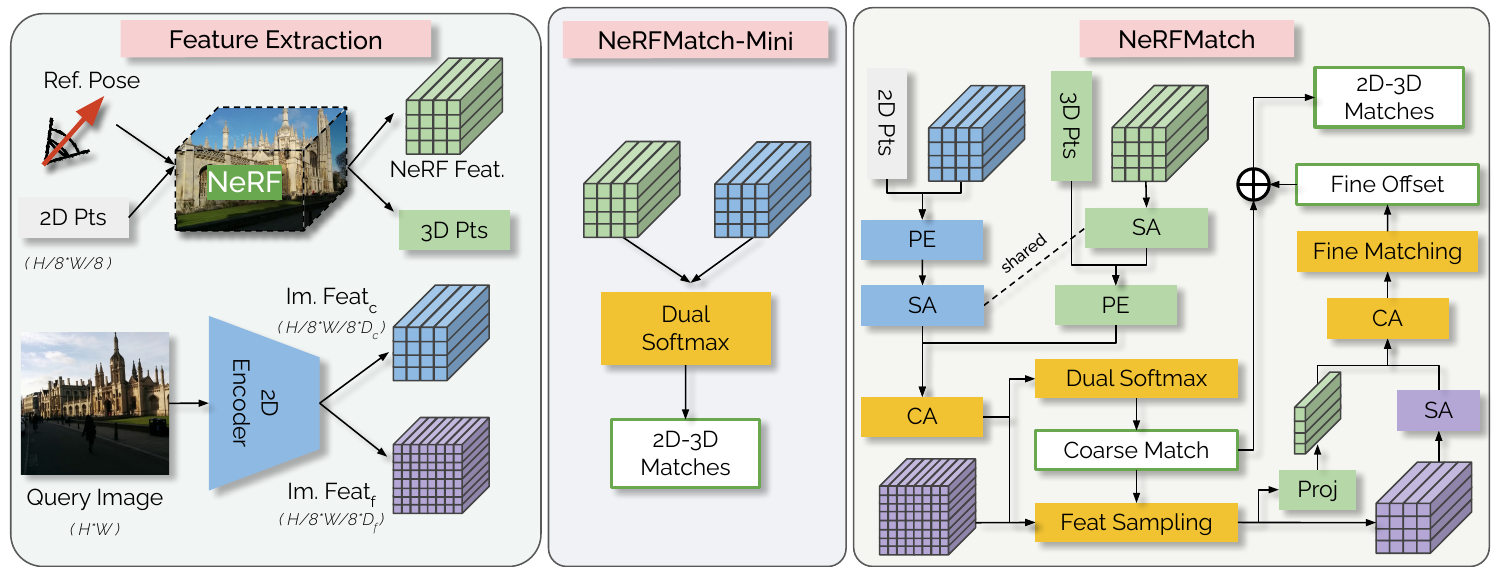}
\end{center}
   \caption{\textbf{\nerfmatch architecture.} We present \nerfmatch as our full matching model (\textit{rightmost}) and \nerfmatch-Mini as a light version of it (\textit{middle}). Both models share the same feature extraction process, where we use a 2D encoder to extract image features at two resolutions and render 3D points with associated \nerf features at sampled 2D pixel locations from the reference viewpoint. 
   The full matching uses self-attention (SA) and cross-attention (CA) with positional encodings (PE).
   }
\label{fig:nerfmatch}
\end{figure}

\PAR{Overview.}
To establish 2D-3D correspondences between a query image and \nerf scene points for localization, we introduce two variants of our proposed image-to-\nerf matching network, a \textit{full} version (\nerfmatch) and a \textit{minimal} version (\nerfmatch-Mini), as depicted in \cref{fig:nerfmatch}.
The \textit{full} version is notably more expressive, incorporating powerful attention modules~\cite{sarlin2020superglue, sun2021loftr, chen2022aspanformer} and follows a coarse-to-fine matching paradigm~\cite{sun2021loftr, zhou2021patch2pix, chen2022aspanformer}. While it delivers more accurate matches it is computationally more expensive (\cf \cref{exp:match_ablat}). Consequently, we also propose a \textit{minimal} version of our method that focuses on learning good features for matching, removing the need to learning the matching function itself. 
Both models comprise a feature extraction module that encodes both the query image and 3D scene points into a feature space, and a matching module that aligns these two feature sets to determine the 2D-3D correspondences. We further detail their architectural designs and supervision in the subsequent subsections.

\subsection{\nerfmatch-Mini}\label{sec:nerfmatch_mini}
\PAR{Image encoding.}
Given a query image $I \in \mathbb{R}^{H \times W \times 3}$, we use a CNN encoder to extract a coarse-level feature map $F_{m}^c \in \mathbb{R}^{N_m \times D^c}$ with $N_{m}=\frac{H}{8} \times \frac{W}{8}$ and a fine-level feature map $F_{m}^f \in \mathbb{R}^{\frac{H}{2} \times \frac{W}{2} \times D^f}$. 

\PAR{\nerf feature encoding.}
Given the reference pose of the query image found by image retrieval (\cf \cref{sec:method_loc_pipeline}), we use \nerf to obtain a set of scene 3D points $X_s \in \mathbb{R}^{N_s \times 3}$ and their associating \nerf features $F_{s} \in \mathbb{R}^{N_s \times D^c}$ (\cf \cref{sec:method_nerf_scene}). 
To make the matching module memory managable, we consider \nerf features $F_{s} \in \mathbb{R}^{N_s \times D^c}$ rendered along rays originated at center pixels of every $8 \times 8$ image patch in the reference view, \ie, $N_{s}=\frac{H}{8} \times \frac{W}{8}$.

\PAR{Dual-softmax matching.} 
After the feature extraction, \nerfmatch-Mini directly matches the \nerf feature map $F_s$ against the coarse-level image feature map $F^c_m$ with a non-learnable dual-softmax matching function adopted from ~\cite{sun2021loftr}. Specifically, we compute pair-wise cosine similarities between the two feature maps followed by a dual-softmax operation to obtain the matching score matrix $S \in  \mathbb{R}^{N_{m} \times N_{s}}$. Finally, we extract mutual matches based on the association scores which gives us the predicted matches.

\subsection{\nerfmatch}\label{sec:nerfmatch}
As depicted in \cref{fig:nerfmatch}, \nerfmatch shares the same feature extraction processes as \nerfmatch-Mini, but learns an attention-based matching module which follows the coarse-to-fine paradigm of image-to-image feature matching~\cite{zhou2021patch2pix, sun2021loftr, chen2022aspanformer}: (i) firstly, we identify 3D-to-image patch matches using a coarse matching module, (ii) secondly, we refine to pixel accuracy with a fine matching module. We describe the detailed matching module architecture in the following paragraphs.

\PAR{Coarse-level matching.}
Equipped with coarse image features $F_m^c$ that represent image local patches and raw \nerf features $F_s$ that represent individual 3D scene points, we first apply 2D positional encoding~\cite{sun2021loftr, carion2020detr} to equip the image features with positional information.
We next enrich the global contextual information within each domain by applying several self-attention blocks. We share the self-attention weights between image features and \nerf features to help bring the features from two different domains into a common embedding space for matching. 
After self-attention, we enhance the 3D information explicitly by concatenating each \nerf feature with its positional-encoded 3D points using \nerf positional encoding~\cite{mildenhall2020nerf}. 
Afterwards, we feed those features into a cross-attention layer to enable cross-domain interactions. 
We use the same dual-softmax matching function introduced above (\cf \cref{sec:nerfmatch_mini}) to obtain coarse matches $\mathcal{M}_c=\{(i, j) |i \in (0, N_m-1), j \in (0, N_s-1)\}$ where $i, j$ are the indices of the image and point features.

\PAR{Fine-level matching.} 
For each coarse match $m_c = (i, j)$ that we extract, we start by gathering its high-resolution image patch feature $F_m^f(i) \in \mathbb{R}^{w \times w \times D^f}$ from the fine-level feature map, centered around the corresponding location of the match. Inside each of these image patches, a self-attention block is applied to spread contextual information throughout the patch.
Next, for every 3D point feature, we take the cross-attended feature obtained from the coarse matching, and process it through a linear layer to adjust its feature dimension from $D^c$ to $D^f$.
In the subsequent step, we align the 3D feature with its corresponding local feature map. This alignment~\cite{sun2021loftr} produces a heatmap, which depicts the likelihood of the 3D point $j$ matching with each pixel in the vicinity of image pixel $i$. To obtain the exact, fine match, we compute the expected value across this heatmap's probability distribution. The final, refined matches obtained through this process are denoted as $\mathcal{M}_f$.

\subsection{Supervision}
\PAR{Ground-truth matches.} 
We project the rendered 3D scene points $X_s$ onto the query image using the ground-truth query camera calibration, which gives their precise 2D projections $x_s$ at the resolution in which we want to supervise the fine matches.
To compute the ground-truth coarse association matrix $M_{gt}$ which is a binary mask, for each 3D point $j$, we assign it to its belonging $i$-th $8 \times 8$ local patch in the query image. We set the association value $M_{gt}(i, j) = 1$ if a 3D point $j$ finds its 2D patch $i$ within the query image boundary. Notice, one image patch can be assigned to multiple 3D points yet each 3D point has at most one 2D match.

\PAR{Losses.}  To supervise the coarse matching, we apply the log loss~\cite{sun2021loftr} to increase the dual-softmax probability at the ground-truth matching locations in $M_{gt}$. The coarse matching loss $L_c$ is defined as:
\begin{equation}\label{eq:coarse_matching_loss}
L_c = - \frac{1}{M_{gt}} \sum_{(i, j) \in \mathcal{M}_{gt}} \log (S(i, j)).
\end{equation}
To compute the fine matching loss, for a 3D point $X_j$ with ground-truth fine match $x_j$, we supervise its predicted fine match $\Tilde{x}_j$ by minimizing its pixel distance to the ground-truth match.
Following ~\cite{wang2020caps, sun2021loftr}, we compute the total variance $\sigma^{2}(j)$ of the corresponding heatmap and minimize the weighted loss function:
\begin{equation}\label{eq:fine_matching_loss}
L_f = \frac{1}{M_{f}} \sum_{(i, j)\in M_f} \frac{1}{\sigma^2(i)} ||\Tilde{x}_j - x_j||_2.
\end{equation}

The \nerfmatch-Mini model is supervised with only the coarse loss $L_c$ while for \nerfmatch model is supervised with the sum of coarse and fine matching losses $L_c + L_f$.

\section{Experiments}
\label{sec:experiments}

\PAR{Datasets.} Cambridge Landmarks~\cite{kendall2015posenet} is a dataset of handheld smartphone images of $6$ outdoor scenes with large exposure variations which is considered challenging for \nerf techniques. We follow previous work~\cite{brachmann2023ace, sarlin2021pixloc} and evaluate on a subset of $5$ scenes whose spatial extent ranges from $875 m^2$ to $5600 m^2$. 
We also test our method on 7-Scenes~\cite{shotton2013scene7}, which is composed of RGB-D images captured in 7 unique indoor scenes whose size ranges from $1 m^3$ to $18 m^3$. Its images contain large texture-less surfaces, motion blur, and object occlusions.
For both datasets, we follow the original released training and testing splits. 
Following recent work~\cite{brachmann2023ace, chen2023nefes, brachmann2021pgt}, we use the more accurate SfM pose annotations for 7-Scenes rather than its original pose annotations.

\PAR{Evaluation metrics.} We report median pose errors, \ie, translation error in centimeters and median rotation error in degrees. We further report localization recall that measures the percentage of queries localized with pose errors below specific thresholds, \ie, $(5 cm , 5^\circ)$ for 7-Scenes. 
As Cambridge Landmarks has large variation in scene scales, we follow ~\cite{brachmann2021dsac*, liu2023nerfloc} to use $5^\circ $ rotation error and variable translation error thresholds, \ie, $38/22/15/35/45 cm$ for  King’s College/Old Hospital/Shop Facade/St. Mary’s Church/Great Court.

\PAR{Implementation details.} We use the first two blocks of ConvFormer~\cite{yu2022metaformer} as the image backbone and initialize it with ImageNet-1K~\cite{russakovsky2015imagenet} pre-trained weights\footnote{The weights can be downloaded from \url{huggingface.co/timm/convformer_b36.sail_in1k_384}}. We set feature dimensions for coarse and fine matching as $D^c=256$ and $D^f=128$. For fine matching, we use local window size $w=5$ for image feature cropping.
We resize query images to $480 \times 480$ for all experiments. We train minimal and full \nerfmatch models using Adam~\cite{kingma2014adam} optimizer with canonical learning rate $clr=0.0008/0.0004$ and batch size $cbs=16$ for $30/50$ epochs accordingly. We decay the learning rate based on the cosine annealing scheduling~\cite{loshchilov2016sgdr}. Our models are trained on $8$ Nvidia V100 GPUs (16/32GB). 
A MipNeRF model is $5.28$ MB in size. Given a camera pose, it takes $141$ milliseconds to render $3600$ 3D points with its features on a single 16GB Nividia V100 GPU.
Our \nerfmatch-Mini and \nerfmatch models are $42.8/50.4$ MB  in size and require $37/157 ms$  to run a forward pass at $480\times480$ image resolution. 
We provide \nerf implementation details in the supplementary material.

\subsection{Localization Evaluation}\label{exp:localization}

\PAR{Baselines.} 
We first compare our proposed \nerfmatch and \nerfmatch-Mini against common visual localization approaches on both indoor and outdoor datasets. 
We split the methods into the \textit{end-to-end} category including APR~\cite{shavit2021mstpn, chen2022dfnet, moreau2022lens, chen2023nefes} and SCR~\cite{brachmann2023ace, brachmann2021dsac*, li2020hierarchical} methods, and the \textit{hierarchical} category~\cite{yang2019sanet, tang2021dsm, tang2023neumap, taira2018inloc, sarlin2019hloc, sarlin2021pixloc, moreau2023crossfire, liu2023nerfloc} where methods rely on an extra image retrieval step to coarsely localize the region in the scene. 
In addition, we specify the underlying scene representation used for localization at \textit{test time}.

\PAR{Inference settings.} 
We use \textit{top-1/10} reference poses for Mini/Full \nerfmatch models  for outdoor and \textit{top-1} for indoor and apply the best pose refinement determined in \cref{exp:pose_refinement}. We provide more  details in supplementary.

\begin{table*}[t!]

  \centering
  \setlength{\tabcolsep}{3pt}
  \caption{\textbf{Outdoor localization on Cambridge Landmarks~\cite{kendall2015posenet}.} We report per-scene median rotation and position errors in $(cm, ^\circ)$  and its average across scenes.}
\resizebox{0.75\textwidth}{!} {  
  \begin{tabular}{cllcccccc}
    \toprule
    \multicolumn{2}{c}{\multirow{2}{*}{Method}} & Scene 
    &\multicolumn{6}{c}{Cambridge Landmarks - Outdoor}  \\
    \cmidrule(r){4-9}    
    & & Repres.
    & Kings & Hospital & Shop & StMary & Court & Avg.Med $\downarrow$ \\
    
\midrule
\multirow{7}{*}{\rotatebox{90}{End-to-End}}
    &MS-Trans.~\cite{shavit2021mstpn} & APR Net.
    & 83/1.5 & 181/2.4 & 86/3.1 &  162/4 & - & - \\
    &DFNet~\cite{chen2022dfnet} & APR Net.
    & 73/2.4 & 200 /3 & 67/2.2 & 137/4 & - & - \\
    &LENS~\cite{moreau2022lens} & APR Net.
    & 33/0.5 & 44/0.9 & 27/1.6 & 53/1.6	& -	& -	\\
    &NeFeS~\cite{chen2023nefes} & APR+NeRF
    & 37/0.6 & 55/0.9 & 14/0.5 & 32/1 & - & - \\
    &DSAC*~\cite{brachmann2021dsac*}  & SCR Net.
    & 15/0.3 & 21/0.4 & 5/0.3 & 13/0.4 & 49/0.3 & 20.6/0.3 \\
    &HACNet~\cite{li2020hierarchical} & SCR Net.
    & 18/0.3 & 19/\bf{0.3} & 6/0.3 & 9/0.3 & 28/0.2 & 16/0.3 \\    
    &ACE~\cite{brachmann2023ace} & SCR Net. 
    & 28/0.4 & 31/0.6 & 5/0.3 & 18/0.6 & 43/0.2& 25/0.4 \\
\midrule                                        
\multirow{11}{*}{\rotatebox{90}{Hierachical}}    
    &SANet~\cite{yang2019sanet} & 3D+RGB     
    & 32/0.5 & 32/0.5 & 10/0.5 & 16/0.6 & 328/2.0 & 83.6/0.8  \\    
    &DSM~\cite{tang2021dsm} & SCR Net.          
    & 19/0.4 & 24/0.4 & 7/0.4 & 12/0.4 & 44/0.2 & 21.2/0.4  \\
    &NeuMap~\cite{tang2023neumap} & SCode+RGB   
    & 14/\bf{0.2} & 19/0.4	& 6/0.3 & 17/0.5 & \bf{6/0.1} & 12.4/0.3 \\
    &InLoc\cite{taira2018inloc} & 3D+RGB    
    & 46/0.8 & 48/1.0 & 11/0.5 & 18/0.6 & 120/0.6 & 48.6/0.7 \\    
    &HLoc\cite{sarlin2019hloc} & 3D+RGB
    &12/\bf{0.2} & \bf{15/0.3} & \bf{4/0.2} & \bf{7/0.2} & 16/\bf{0.1} & 10.8/\bf{0.2} \\
    &PixLoc\cite{sarlin2021pixloc} & 3D+RGB
    & 14/\bf{0.2} & 16/\bf{0.3} & 5/\bf{0.2} &10/0.3 & 30/\bf{0.1} & 15/\bf{0.2} \\
    &CrossFire~\cite{moreau2023crossfire} & NeRF+RGB
    & 47/0.7 & 43/0.7 & 20/1.2 & 39/1.4 & - & - \\
    &NeRFLoc~\cite{liu2023nerfloc} & NeRF+RGBD
    & \bf{11/0.2} & 18/0.4 & \bf{4/0.2} & \bf{7/0.2} & 25/\bf{0.1} & 13/\bf{0.2} \\

    \cmidrule(r){4-9}
    &\nerfmatch-Mini & NeRF+RGB
    & 19.0/0.3	& 30.2/0.6	& 10.3/0.5	& 11.3/0.4 & 29.1/0.2 & 20.0/0.4 \\
    &\nerfmatch & NeRF+RGB    
    & 13.0/\bf{0.2}	& 19.4/0.4 & 8.5/0.4 & 7.9/0.3 & 17.5/\bf{0.1}	& 13.3/0.3 \\

    &\nerfmatch & NeRF 
    & 12.7/\bf{0.2} & 20.7/0.4 & 8.7/0.4 & 11.3/0.4 & 19.5/\bf{0.1} & 14.6/0.3 \\
    \bottomrule
  \end{tabular}
  }
    \label{tab:loc_camb}
\end{table*}

\PAR{Results on Cambridge Landmarks.} As shown in \cref{tab:loc_camb}, our \textit{minimal} version despite being lightweight, is able to achieve comparative results \wrt most of the SCR methods and surpass all APR methods. With a more advanced attention-based hierarchical matching function, our \textit{full} model achieves the competitive results among all methods. Our experiments fully demonstrate that the \nerf inner features learned via view synthesis are discriminative 3D representations for 2D-3D matching.

\begin{table*}[t!]

  \centering
  \setlength{\tabcolsep}{3pt}
  \caption{\textbf{Indoor localization on 7-Scenes~\cite{shotton2013scene7}.}  We report per-scene median rotation and position errors in $(cm, ^\circ)$  and their average across scenes, along with averaged localization recall.  }
\resizebox{0.9\textwidth}{!} {  
  \begin{tabular}{llccccccccc}
    \toprule
    \multirow{2}{*}{Method} & Scene &\multicolumn{9}{c}{7-Scenes - SfM Poses - Indoor} \\
    \cmidrule(r){3-11}
    & Repres. & Chess & Fire & Heads & Office & Pump. & Kitchen & Stairs &  Avg.Med$\downarrow$ & Avg.Recall$\uparrow$.\\
    
\midrule
    MS-Trans.~\cite{shavit2021mstpn} & APR Net.    
    & 11/6.4 & 23/11.5 & 13/13 & 18/8.1 & 17/8.4 & 16/8.9 & 29/10.3 & 18.1/9.5 & - \\
    DFNet~\cite{chen2022dfnet} & APR Net.
    & 3/1.1 & 6/2.3 & 4/2.3 & 6/1.5 & 7/1.9 & 7/1.7 & 12/2.6  & 6.4/1.9 & -\\
    NeFeS~\cite{chen2023nefes} & APR+NeRF
    & 2/0.8 & 2/0.8 & 2/1.4 & 2/0.6 & 2/0.6 & 2/0.6 & 5/1.3 & 2.4/0.9 & - \\
    \cmidrule(r){3-11}
    DSAC*~\cite{brachmann2021dsac*} & SCR Net. 
    & 0.5/0.2 & 0.8/0.3	& 0.5/0.3 & 1.2/0.3	& 1.2/0.3 & 0.7/0.2 & 2.7/0.8 & 1.1/0.3 & \bf{97.8} \\
    ACE~\cite{brachmann2023ace} & SCR Net. 
    &0.7/0.5 & 0.6/0.9 & 0.5/ 0.5 & 1.2/0.5	& 1.1/0.2 & 0.9/0.5	& 2.8/1.0 & 1.1/0.6 & 97.1 \\
    DVLAD+R2D2\cite{torii2015densevlad, revaud2019r2d2} & 3D+RGB 
    &  \bf{0.4/0.1} & \bf{0.5/0.2} & \bf{0.4/0.2} & \bf{0.7/0.2} & \bf{0.6/0.1} & \bf{0.4/0.1} & \bf{2.4/0.7} & \bf{0.8/0.2} & 95.7 \\
    HLoc\cite{sarlin2019hloc} & 3D+RGB    
    & 0.8/\bf{0.1} & 0.9/\bf{0.2}	& 0.6/0.3 & 1.2/\bf{0.2} & 1.4/0.2 & 1.1/\textbf{0.1}	& 2.9/0.8 & 1.3/0.3 & 95.7  \\
    \cmidrule(r){3-11}
    \nerfmatch-Mini & NeRF+RGB
    & 1.6/0.5 &	1.5/0.6 & 1.4/0.9 &	3.6/1.0 & 3.5/0.9 & 1.7/0.5 & 8.5/2.1 &	3.1/0.9 & 74.4 \\
    \nerfmatch & NeRF+RGB    
    &  0.9/0.3 & 1.1/0.4 & 1.4/1.0 & 3.0/0.8 & 2.2/0.6 & 1.0/0.3 & 9.0/1.5 & 2.7/0.7 & 78.2 \\

    \nerfmatch & NeRF 
    &  0.9/0.3 & 1.1/0.4 & 1.5/1.0 & 3.0/0.8 & 2.2/0.6 & 1.0/0.3 & 10.1/1.7 & 2.8/0.7 & 78.4 \\
    \bottomrule
  \end{tabular}
  }
    \label{tab:loc_7sc}
\end{table*}

\PAR{Results on 7-Scenes.} As shown in \cref{tab:loc_7sc}, we are on-par with the best APR method (the \textit{upper} rows), NeFeS~\cite{chen2023nefes}, while we are less accurate than SCR and visual matching methods whose accuracy on 7-Scenes is close to saturation (the \textit{middle} rows). 
Our hypothesis is that those method are able to benefit from the dense distribution of frames in the sequence. However, we have to limit our \nerf training to 900 training frames (loaded at once into memory) per scene for efficient training despite thousands of frames are available.

\PAR{\nerf-only localization.} 
We further push our method to using \nerf as the \textit{only} scene representation for localization, which means we would no longer need access to a real image database as part of the scene representation, which requires a significantly bigger storage than a single \nerf model.
For this purpose, we propose to perform image retrieval on synthesized images rendered by our \nerf model. This entails a small decrease in performance on Cambridge in translation error. 
This slight degradation in performance can be attributed to the increased complexity of these scenes for \nerf. This assertion is supported by the observation that, when switching to image retrieval on synthesized images, our model demonstrates almost no change in performance on the indoor 7-Scenes dataset. Our results open the door to a localization pipeline where only a \nerf model is needed.

In the following experiments, we conduct ablation studies to thoroughly understand the different components of our method. We conduct all ablations on Cambridge Landmarks.

\subsection{\nerf Feature Ablation}\label{exp:nerf_ablat}
\begin{table}[t!]
    \centering
    \setlength{\tabcolsep}{10pt}
    \caption{ \textbf{\nerf feature ablation on Cambridge~\cite{kendall2015posenet}.}  We train \nerfmatch-Mini with different 3D features and compare their localization performance. }
    \resizebox{0.9\textwidth}{!}{%
        \begin{tabular}{lccccccccc} 
         \toprule            
             Metrics & 
             Pt3D & Pe3D & $f^1$ & $f^2$ & $f^3$ & $f^4$ & $f^5$ & $f^6$ & $f^7$\\ 
             \midrule             

             Med. Translation ($cm,\downarrow$) &  
             458.0 & 34.3 & 28.7 & 28.4 & \textbf{27.9} & 28.3 & 28.3 & 30.2 & 61.3 \\
             Med. Rotation ($^\circ, \downarrow$) &
             6.5 & 0.6 & \textbf{0.5} & \textbf{0.5} &  \textbf{0.5} & \textbf{0.5} & \textbf{0.5} & \textbf{0.5} & 1.3 \\
             Localize Recall. ($\%,\uparrow$) & 
             0.7 & 51.4 & 58.6 & \textbf{59.4} & \textbf{59.2} & 56.9 & 57.7 & 53.0 & 38.8\\
        \bottomrule
        \end{tabular}
    }
    
    \label{tab:nerf_feat_ablat}
\end{table}

\PAR{Experimental setup.}
As our first ablation, we investigate the potential of \nerf features in 2D-3D matching. We consider several different types of 3D features including the raw 3D point coordinates (Pt3D), positional encoded 3D points (Pe3D), and \nerf inner features output from  all intermediate layers $f^{j}$ with $j \in [1, 7]$, as shown in \cref{fig:nerf_arch}.
\nerf features have the same dimension as the image backbone features, and thus are directly ready for 2D-3D matching. 
For 3D point coordinates and positional encoded 3D points as 3D features, we use a simple linear layer to lift them to the image feature dimension, and we train them together with the image backbone on feature matching.
To fully focus on the influence of different 3D features, we train \nerfmatch-Mini models with different 3D features and conduct matching without pose refinement.
We evaluate the quality of 2D-3D matches for localization and report the median pose errors and localization recall in \cref{tab:nerf_feat_ablat}.

\PAR{Results.} We show that directly matching image features with lifted 3D coordinate features does not yield accurate results. This significantly improves when using a \nerf positional encoding layer~\cite{mildenhall2020nerf} on top of the raw 3D coordinates. Notably, the Pe3D feature corresponds to the input for \nerf. 
We further show that all first six layer \nerf features are better than Pe3D in aligning image features to 3D features, showing that matching benefits from the view synthesis training. 
The only exception is that the last layer feature $f^7$, which produces features that are less discriminative for matching and are more focused on the \nerf goal of  predicting density and RGB values.
Among layer features, we found the middle layer $f^3$ to be slightly better than the others and choose it to be the default \nerf feature we use for our \nerfmatch models.

\subsection{\nerfmatch Ablation} \label{exp:match_ablat}

\PAR{Architecture ablation.}
Next, we study the influence of different image backbones and matching functions on our matching model in \cref{tab:matcher_ablation}.  
We first compare two convolution backbones for image feature encoding, \ie, ResNet34~\cite{he2016resnet} and ConvFormer~\cite{yu2022metaformer}.
Our method demonstrates improved performance with the latter, which is consistent with their performance on image classification~\cite{russakovsky2015imagenet}.  
We also observe that large-scale ImageNet~\cite{russakovsky2015imagenet} pre-training provides a better starting point for extracting suitable image features for matching, leading to increased localization accuracy compared to training from scratch. 
We further show that for the same backbone, a more advanced attention matching function with a coarse-to-fine design is crucial for accurate matching and significantly improves localization accuracy, albeit at the cost of increased runtime. 

\begin{table}[t!]
    \centering
    \setlength{\tabcolsep}{3pt}
    \caption{ \textbf{\nerfmatch architecture ablation on Cambridge Landmarks~\cite{kendall2015posenet}.} We report averaged median pose error in $(cm, ^\circ)$ and localization recall. }
    \resizebox{0.88\textwidth}{!}{%
        \begin{tabular}{lcccccccc}     
         \toprule            
             Model  & \multirow{2}{*}{Backbone}  & Pretrain & \multirow{2}{*}{Matcher}  & Training & Avg.Med & Avg.Recall & 
 Model Size & Runtime \\ 
              Name & & Backbone & & Scenes & $(cm/^\circ)\downarrow$ & $(\%)\uparrow$ & $(MB)\downarrow$ & $(ms)\downarrow$ \\
             \midrule
    
             - & ResNet34   & \cmark & Minimal & Per-Scene   & 32.7/0.6 & 52.0 & \textbf{32.8} & \textbf{23}\\
             
             - & ConvFormer & \xmark & Minimal & Per-Scene   & 34.8/0.6 & 50.7 & 42.8 & 37\\
             \nerfmatch-Mini  & ConvFormer & \cmark & Minimal & Per-Scene   & 27.9/0.5 & 59.2 & 42.8 & 37\\
             \nerfmatch-Mini (MS) & ConvFormer & \cmark & Minimal & Multi-Scene & 30.8/0.5 & 53.6 & 42.8 & 37\\
             \nerfmatch & ConvFormer & \cmark & Full & Per-Scene   & \textbf{16.5/0.3} & \textbf{71.3} & 50.4 & 157 \\
             \nerfmatch (MS)  & ConvFormer & \cmark & Full & Multi-Scene & 22.0/0.4 & 65.2 & 50.4 & 157 \\
             \bottomrule
        \end{tabular}
    }    
    \label{tab:matcher_ablation}
\end{table}

\PAR{Training ablation.}
In addition to architecture choices, we also examine the influence of different training settings, \ie, training per-scene (default) and training multi-scenes (all scenes within a dataset).  
Despite \nerf features being trained per-scene, we surprisingly find that both minimal and full \nerfmatch models can be trained to handle multi-scene (MS) localization with only a slight decrease in accuracy \wrt the per-scene model. 
This is a similar finding as recent scene-agnostic SCR methods~\cite{yang2019sanet, tang2023neumap, tang2021dsm}, that extend per-scene SCR to multi-scenes by conditioning SCR on scene-specific 3D points.
While scene-agnostic SCR learns to regress directly the 3D features in the form of xyz coordinates, our models learn to find a common ground between image features and \nerf features for matching.

\subsection{Pose Refinement} \label{exp:pose_refinement}

\begin{figure}[t]
\begin{minipage}{0.5\linewidth}
    \includegraphics[width=\linewidth]{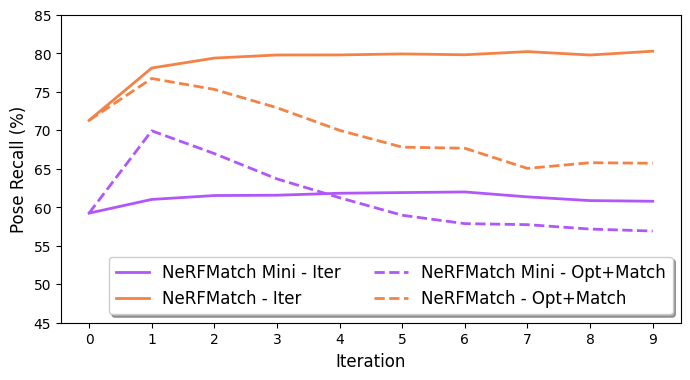}        
\end{minipage}
\begin{minipage}{\linewidth}
\resizebox{0.5\linewidth}{!} {        
    \setlength{\tabcolsep}{3pt}

    \begin{tabular}{lcccc}
    \toprule

    \multirow{2}{*}{Model} & Best  & No Refinement & Refined & Refined\\
        &   Refinement & (top$-1$) & (top$-1$) & (top$-10$) \\
  \multicolumn{2}{l}{Metrics} &\multicolumn{3}{c}{ Avg.Med $(cm/^\circ)\downarrow$ / Avg.Recall $(\%)\uparrow$} \\
    
    \midrule    
    \nerfmatch-Mini & Opt+Match & 27.9/0.5/59.2 &  20.5/0.4/70.9 & 20.5/0.4/70.9\\
    
    \nerfmatch &  Iter. & 16.5/0.3/71.3 &  14.2/0.3/78.2 & 13.3/0.3/80.8\\

    \bottomrule
    \end{tabular}
}   
\end{minipage}

\caption{
\textbf{Refinement ablation on Cambridge Landmarks~\cite{kendall2015posenet}.}
On the left side, we depict the average recall for optimization-based (Opt+Match) and iterative (Iter) refinement approaches across multiple iterations. 
We provide results for both the \nerfmatch and its minimal setting. 
On the right side, we report averaged median pose error in ($cm/^\circ$) and localization recall with the best refinement configurations.
}
\label{fig:refinement}
\end{figure}

After conducting matching architecture ablation, we investigate different refinement methods to further increase pose accuracy.
We examine two approaches: (i) an iterative approach, where we re-run matching with the last computed estimation as reference pose, and (ii) an optimization approach, where we optimize a reference pose through frozen NeRF weights using a photometric loss and then run the final matching. 
As shown in \cref{fig:refinement}, both methods show an improvement over the initial estimate. We assess both the NeRFMatch model and its minimal version (Mini) trained per-scene.

\PAR{Iterative refinement for computational efficiency.} 
In the NeRFMatch setting, both refinement approaches show similar early results, since the initial pose estimate is relatively close to the solution, but the iterative approach is more stable over time.
Additionally, it is worth noting that the optimization approach incurs a higher computational cost compared to the iterative approach.
The runtime for the optimization refinement excluding the matching step (shown in \cref{tab:matcher_ablation}) is $398.4 ms$ for a single optimization step and subsequent rendering. In contrast, for a single step of iterative refinement, the runtime amounts to $141.2 ms$.
Given the better performance and the quicker computational time, we opt for the iterative approach as the default refinement for NeRFMatch.

\PAR{Optimization refinement for large pose corrections.} 
In comparison to the iterative method, the optimization approach significantly benefits the minimal matching model. 
 This is because the initial query pose estimation is notably distant compared to the one given by NeRFMatch, and subsequent iterations only result in incremental refinements.
 The optimization approach takes a more substantial step towards the query pose, and achieving a more optimal reference pose results in improved query pose estimation. 
 We have already seen that the optimization refinement incurs higher computational time, but it also depends on a learning rate schedule, which is sensitive to configuration.
Through empirical analysis, we select $1\times10^{-3}$ as the initial learning rate and apply a cosine annealing learning rate schedule.
 We set the optimization approach as the default refinement for NeRFMatch-Mini.

\section{Conclusion and Limitations}
\label{sec:conclusion}
In this work, we have taken initial steps towards leveraging \nerf as the primary representation for the task of camera localization. To achieve this, we have thoroughly examined the performance of \nerf features in the localization task, considering various architectural designs, feature extraction from different encoder layers, and diverse training configurations. Additionally, we have demonstrated that \nerf can remove the need for the original image set for coarse localization. 
Our results suggest that \nerf features are highly effective for 2D-3D matching.

While \nerfmatch marks a significant step towards comprehensive localization using \nerf, it also highlights several limitations that necessitate further research. Specifically, 
we observe a noticeable performance gap when applying our method to indoor 7-Scenes dataset. 
  
\title{NeRFMatch | Supplementary}
\author{}
\institute{}
\maketitle

\appendix
In this supplementary document, we provide further details regarding our proposed method and qualitative results. 
We describe our implementation details for \nerf in \cref{sec:supp_nerf_impl}, and for \nerfmatch in \cref{sec:supp_nerfmatch_impl}. 
Then, we present additional analysis and discussion of our method in \cref{sec:supp_analysis}.

\begin{figure}[h!]
\begin{center}
\includegraphics[width=0.99\linewidth, trim={0px 0px 0px 0px}, clip]{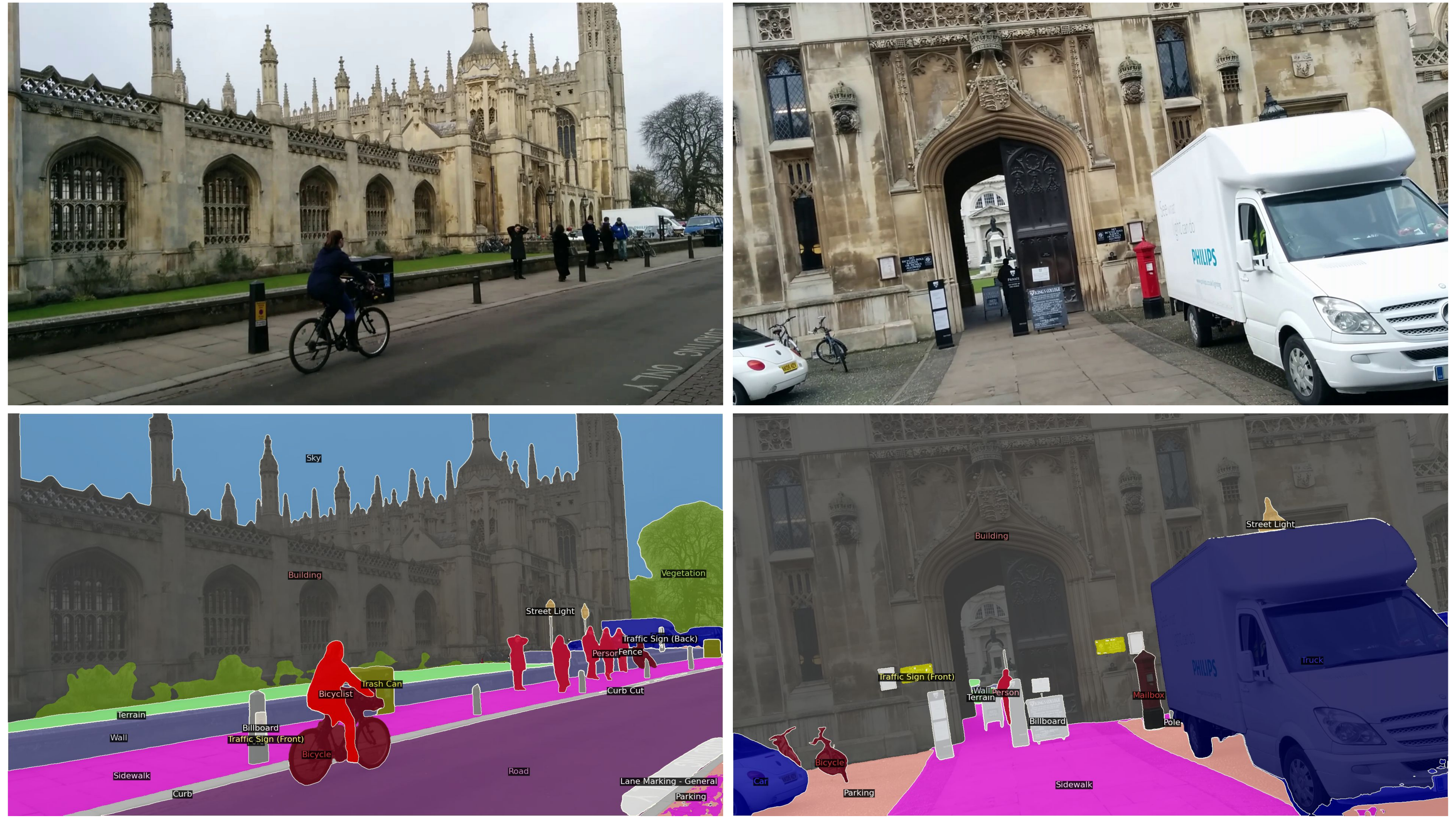}
\end{center}
   \caption{
   Example of masking on Kings College scene. Top images - original images, bottom - semantic segmentation using~\cite{cheng2022masked}.
   }
\label{fig:masking}
\end{figure}

\section{\nerf Implementation Details} \label{sec:supp_nerf_impl}
\PAR{Handling challenges in outdoor scenes.}  
Outdoor reconstruction in the wild has a lot of challenges including illumination changes, transient objects, and distant regions. %
For the task of localization, we are interested in reconstructing only the static scene elements, \eg, roads, buildings, and signs.

\begin{figure*}[t!]
\begin{center}
\includegraphics[width=0.95\linewidth, trim={0px 450px 0px 0px}, clip]{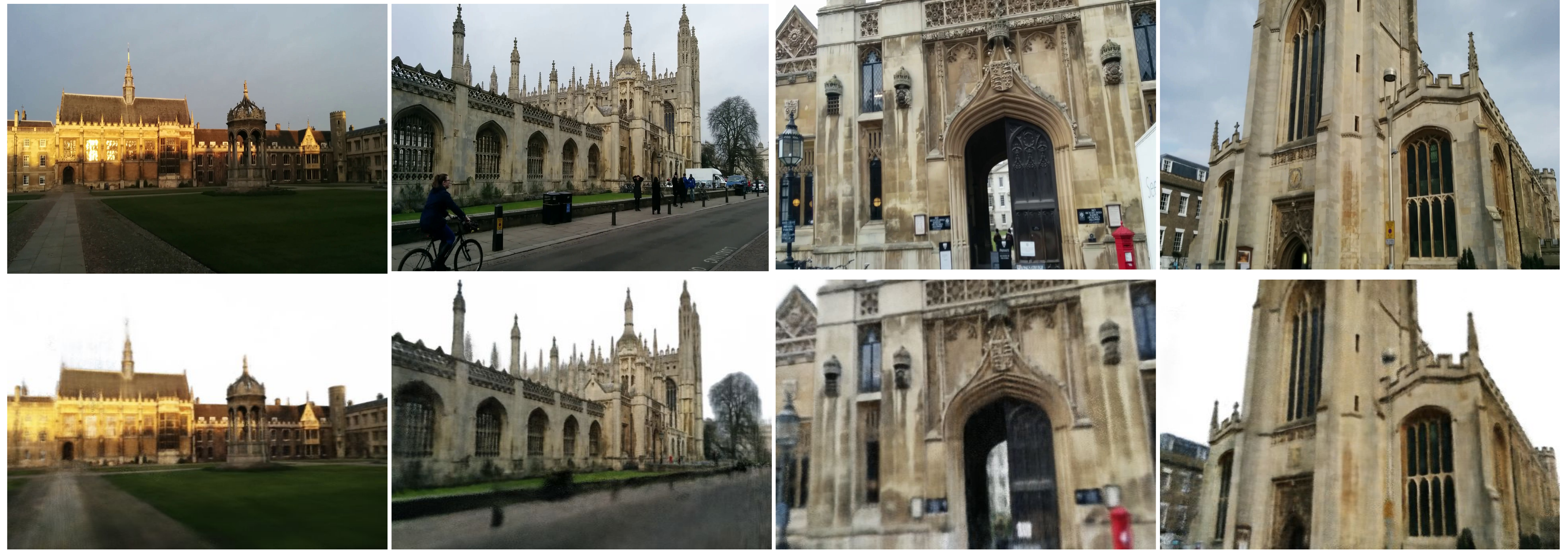}
\end{center}
   \caption{
   Example of masking on the King's College scene of Cambridge Landmarks~\cite{kendall2015posenet}. The bottom row are rendered with NeRF, and the top row - ground truth images.
   }
\label{fig:nerf_masking_effect}
\end{figure*}

To properly train \nerf in such a scenario, we use a pre-trained semantic segmentation model~\cite{cheng2022masked} and mask out the sky and transient objects: pedestrians, bicycles, and vehicles. These objects occupy only a minor part of the captured images and are excluded from the loss computation during the training process. 
Analogous methods for masking in sky regions and/or dynamic object areas have been implemented in other works focused on the reconstruction of urban scenes \cite{tancik2022blocknerf, rematas2022urf, xie2023s}.
We show examples of semantic segmentation in~\cref{fig:masking} and its effect on synthesized views in 
\cref{fig:nerf_masking_effect}.

To account for illumination changes, we use an appearance vector that we concatenate together with the view direction as input, similar to~\cite{martin2021nerfw}. The appearance vector changes across sequences but stays the same for all frames in one sequence since appearance does not drastically change inside a sequence.

\PAR{NeRF architecture.} 
Our NeRF model consists of a MipNeRF~\cite{barron2021mipnerf} architecture with both coarse and fine networks. We utilize the final outputs from the fine network to render RGB, depth maps, and 3D features.

\PAR{\nerf training.}
For each scene, we load a subset of 900 training images and 8 validation images and train each model for $15$ epochs. 
From the set of all pixels in all training samples, we randomly sample a batch of $9216$ rays. Subsequently, for each ray, we sample $128$ points for the coarse network and an additional $128$ for the fine network. 
We use the Adam optimizer~\cite{kingma2014adam} with a learning rate $1.6 \times 10^{-3}$ and cosine annealing schedule~\cite{loshchilov2016sgdr}.
In \cref{tab:nerf_psnr}, we present the per-scene PSNR scores for our trained models on the training images.

\begin{table*}[t!]

  \centering
  \setlength{\tabcolsep}{3pt}
  \caption{\textbf{NeRF PSNR scores.} We present the PSNR scores for our trained MipNeRF models on each scene of Cambridge Landmarks~\cite{kendall2015posenet} and 7-Scenes~\cite{shotton2013scene7}.}
\resizebox{0.98\textwidth}{!} {  
  \begin{tabular}{cccccc|cccccccc}
    \toprule
    \multicolumn{6}{c|}{Cambridge Landmarks - Outdoor} & \multicolumn{8}{c}{7-Scenes - indoor}                 \\
    Kings & Hospital & Shop & StMary & Court & Average 
    & Chess & Fire & Heads & Office & Pump. & Kitchen & Stairs & Average\\
  \midrule    
    22.9	& 22.1 & 24.0 & 23.0 & 23.2 & 23.1  & 29.6 & 30.0 & 32.5 & 30.2 & 31.4 & 27.9 & 34.7 & 30.9 \\
 \bottomrule
  \end{tabular}
  }
    \label{tab:nerf_psnr}
\end{table*}

\section{\nerfmatch Implementation Details} \label{sec:supp_nerfmatch_impl}
We summarize average runtime performance for NeRF and both matching models in \cref{tab:runtime}.

\begin{table}[t!]
    \centering
    \setlength{\tabcolsep}{10pt}
    \caption{\textbf{Runtime.}  
    We show runtime of \nerfmatch-Mini and \nerfmatch. For pose refinement we are using optimization refinement for \nerfmatch-Mini and iterative refinement for \nerfmatch.
    }
    \resizebox{0.7\textwidth}{!}{%
        \begin{tabular}{lcc} 
         \toprule                                   
           
             NeRF type & \nerfmatch-Mini & \nerfmatch\\
              
             \midrule                          
             NeRF render  &  141ms & 141ms \\
             
             Image-to-NeRF matching & 37ms & 157ms\\

             Pose refinement &  398ms & 141ms\\

         \bottomrule
        \end{tabular}
    }
    \label{tab:runtime}
\end{table}

\PAR{Training pairs.} We use the same training pairs\footnote{Image pairs are available from \url{https://cvg-data.inf.ethz.ch/pixloc_CVPR2021/}} generated by PixLoc\cite{sarlin2021pixloc} which were computed based on image covisibility within the training split. During training, for each train image we load its top-$20$ covisible pairs. For each training epoch, we then randomly sample $10000$ training pairs from those covisible pairs for each scene. In the case, we train multiple scenes, we merge those pairs across scenes which allows us to balance the training samples across different scenes.

\PAR{Image retrieval.} We adopt the retrieval pairs pre-computed by PixLoc~\cite{sarlin2021pixloc} using NetVLAD~\cite{arandjelovic2016netvlad} for Cambridge Landmarks~\cite{kendall2015posenet} and DenseVLAD~\cite{torii2015densevlad} for 7 Scenes~\cite{shotton2013scene7} during inference. 
We use those retrieval pairs for all experiments by default except for the \nerf-only localization experiment in Sec. 5.1. That experiment is to confirm the feasibility of \nerf-only localization, therefore we run NetVLAD~\cite{arandjelovic2016netvlad} to extract retrieval pairs at image resolution $480 \times 480$  between the real query images and the training images synthesized by \nerf.

During inference, we noticed applying top$-k$ retrieval pairs with $k > 1$ show  evident improvement for \nerfmatch on Cambridge Landmarks. Thus, we set $k=10$ following the common localization practice~\cite{sarlin2021pixloc, sarlin2019hloc}.
For \nerfmatch-Mini, setting $k>1$ did not change much the performance. We suspect this is due to its less accurate matches, which makes the outlier rejection harder when merging noisy correspondences from more pairs. For the indoor 7 Scenes dataset, we use $k=1$ which is sufficient for  relatively small-size scenes. 

\PAR{Optimization refinement.}
Similar to iNeRF~\cite{yen2021inerf}, we are doing a forward pass through frozen NeRF MLP layers using an estimated pose as the initial camera pose. 
Instead of rendering the entire image, we sample and render $3600$ rays, which are equally spread in a grid structure across the image plane. 
The we apply a regular photometric loss between the query image and the rendered image and backpropagate to update the initial camera pose. 
Instead of using the raw updated camera pose, we render the \nerf features and match them with the \nerfmatch to obtain the final camera pose.

\section{Additional Details}\label{sec:supp_analysis}

\PAR{NeRF backbones.} 
In this section, we evaluate additional NeRF type - Instant NGP~\cite{muller2022instantnerf} in comparison to MipNeRF~\cite{barron2021mipnerf}. 
We use MipNeRF for our experiments in the main paper .
As shown in \cref{tab:nerf_arch}, Instant NGP performs significantly worse. 
We hypothesize that this is due to noisy depth reconstruction that is typical for Instant NGP.

N\begin{table}[t!]
    \centering
    \setlength{\tabcolsep}{10pt}
    \caption{\textbf{NeRF backbone ablation on Cambridge Landmarks.}  
    We compare \nerfmatch-Mini and \nerfmatch performances using Instant NGP.
    }
    \resizebox{0.7\textwidth}{!}{%
        \begin{tabular}{lcc} 
         \toprule                                   
           
             \multirow{2}{*}{NeRF type} & \multicolumn{2}{c}{Avg. Med $(cm/^\circ)\downarrow$/Recall $(\%)\uparrow$ } \\                        
              & NeRFMatch-Mini & NeRFMatch  \\
              
             \midrule                          
             Instant NGP & 41.1/0.7/44.4 & 28.1/0.5/61.3\\

             MipNeRF &  20.0/0.4/69.7 & 13.3/0.3/80.8\\
         \bottomrule
        \end{tabular}
    }
    \label{tab:nerf_arch}
\end{table}

\PAR{Impact of scene sizes.}
Scene size affects both NeRF and localization performance, often coupled with scene content and camera pose distribution. 
Ranking scenes by localization errors (lower is better) leads to  OldHospital ($50 \times 40 m^2$)  $>$ KingsCollege ($140\times 40 m^2$) $>$  ShopFacade ($35 \times 25 m^2$) for outdoor and stairs ($2.5 \times 2 \times 1.5 m^3$) $>$ pumpkin ($2.5 \times 2 \times 1 m^3$) $>$  redkitchen ($4\times 4 \times 1.5 m^3$) $>$ chess ($3\times 2 \times 1 m^3$) for indoor. 
This suggests that smaller scenes (OldHospital, stairs) can be more challenging than larger scenes (KingsCollege, redkitchen) due to challenging contents like repetitive structures and texture-less regions.

\PAR{Image retrieval on synthesized views.}
The goal of NeRF-only experiment is to verify the possibility to use NeRF as the only scene representation removing the need to maintain the original image collection. 
Our experiments show a slight performance decrease due to the domain gap between rendered and real images.
Yet, we did not claim an efficient solution for online image retrieval and NeRF rendering. 
Future research is needed to improve its runtime efficiency either via caching scene reference poses in a hierarchical tree structure to fasten the searching process or leveraging any available prior information such as GPS coordinates to quickly find a subset of poses.

\PAR{Indoor performance bottleneck.} 
NeRF predicted depth maps are used to compute pseudo ground-truth for matching supervision. 
Incorrect depth predictions can lead to misaligned feature correspondences. 
In contrast, image matching, SCR, and APR methods use more accurate labels like Colmap camera poses or 3D maps. 
For small-scale indoor scenes, precise supervision is essential to achieve centimeter-level errors. 
Our method based on feature matching, however, scales better than regression-based approaches in larger outdoor scenes. 
Introducing uncertainty measures to ignore inaccurate matches, as in ~\cite{chen2023nerfscr}, and improved NeRF reconstructions with accurate depth maps will benefit our method.

\bibliographystyle{splncs04}
\bibliography{main}
\end{document}